%% file: conference_101719.tex
\documentclass[conference]{IEEEtran}
\IEEEoverridecommandlockouts
\usepackage{array}
\usepackage[caption=false,font=footnotesize,labelfont=sf,textfont=sf]{subfig}
\usepackage{amsthm}
\usepackage{color}
\usepackage{booktabs}
\usepackage{algorithm, algorithmicx}
\usepackage[noend]{algpseudocode}

\usepackage{float}
\usepackage{diagbox}
\usepackage{array} 

\usepackage{multirow}
\usepackage{multicol}
\usepackage{soul}
\usepackage{bm}
\usepackage{marvosym}
\usepackage{url}
\usepackage{graphicx}
\usepackage{subfloat}
\usepackage{caption}
\usepackage{float}
\usepackage{flushend}
\usepackage{enumitem}
\usepackage{hyperref}

\newcommand{\myfw}{\textit{EneAD}}
\newcommand{\tabincell}[2]{\begin{tabular}{@{}#1@{}}#2\end{tabular}}
\usepackage{cite}
\usepackage{amsmath,amssymb,amsfonts}
\usepackage{graphicx}
\usepackage{textcomp}
\usepackage{xcolor}

\def\BibTeX{{\rm B\kern-.05em{\sc i\kern-.025em b}\kern-.08em
    T\kern-.1667em\lower.7ex\hbox{E}\kern-.125emX}}
\makeatletter
\newcommand{\linebreakand}{%
  \end{@IEEEauthorhalign}
  \hfill\mbox{}\par
  \mbox{}\hfill\begin{@IEEEauthorhalign}
}
\makeatother
\begin{document}

\title{Energy-Efficient Autonomous Driving with Adaptive Perception and Robust Decision}


\author{Yuyang~Xia\textsuperscript{1}, Zibo~Liang\textsuperscript{1}, Liwei~Deng\textsuperscript{2}, Yan~Zhao\textsuperscript{3}, Han~Su\textsuperscript{4}\textsuperscript{\Letter}\thanks{\textsuperscript{\Letter}Corresponding authors: Han Su and Kai Zheng.}, Kai~Zheng\textsuperscript{1}\textsuperscript{\Letter}\\
\textsuperscript{1}University of Electronic Science and Technology of China, China \quad \\
\textsuperscript{2}Aalborg University, Denmark\\
\textsuperscript{3}Shenzhen Institute for Advanced Study, University of Electronic Science and Technology of China, China\\
\textsuperscript{4}Yangtze Delta Region Institute(Quzhou), University of Electronic Science and Technology of China, China \\
\{xiayuyang,zbliang\}@std.uestc.edu.cn, lide@cs.aau.dk, \{zhaoyan,hansu,zhengkai\}@uestc.edu.cn
}

\maketitle

\begin{abstract}
Autonomous driving is an emerging technology that is expected to bring significant social, economic, and environmental benefits. However, these benefits come with rising energy consumption by computation engines, limiting the driving range of vehicles, especially electric ones. Perception computing is typically the most power-intensive component, as it relies on large-scale deep learning models to extract environmental features. Recently, numerous studies have employed model compression techniques, such as sparsification, quantization, and distillation, to reduce computational consumption. However, these methods often result in either a substantial model size or a significant drop in perception accuracy compared to high-computation models. To address these challenges, we propose an energy-efficient autonomous driving framework, called \myfw, which includes an adaptive perception and a robust decision module. In the adaptive perception module, a perception optimization strategy is designed from the perspective of data management and tuning. Firstly, we manage multiple perception models with different computational consumption and adjust the execution framerate dynamically. Then, we define them as knobs and design a transferable tuning method based on Bayesian optimization to identify promising knob values that achieve low computation while maintaining desired accuracy. To adaptively switch the knob values in various traffic scenarios, a lightweight classification model is proposed to distinguish the perception difficulty in different scenarios. In the robust decision module, we propose a decision model based on reinforcement learning and design a regularization term to enhance driving stability in the face of perturbed perception results. Extensive experiments evidence the superiority of our framework in both energy consumption and driving performance. \myfw\ can reduce perception consumption by $1.9\times$ to $3.5\times$ and thus improve driving range by $3.9\%$ to $8.5\%$.
\end{abstract}

\begin{IEEEkeywords}
Autonomous Driving, Data Management, Configuration Tuning
\end{IEEEkeywords}

\input{tex/introduction}
\input{tex/related}

\input{tex/problem}
\input{tex/adaptive_perception}

\input{tex/robust_decision}
\input{tex/experiment}

\input{tex/conclusion}
\input{tex/ack}

\bibliographystyle{IEEEtran}
\newpage
\bibliography{IEEEabrv,sample}

\end{document}

%% file: tex/introduction.tex
\section{Introduction}
Autonomous driving has gained broad attention from the public during the last few years~\cite{brenner2018overview,schrank2021urban}. With intelligence, the autonomous vehicle can have a more comprehensive perception of the surrounding traffic environment and make more reasonable driving decisions compared to human drivers. As a result, it is expected to bring society a large number of benefits, including improved mobility and a significant reduction in collisions. However, these benefits come with increased energy consumption of the computation platform, ranging from several hundred watts (W) to over 1 kW of power~\cite{lin2018architectural}. For example, the computing platform using the Nvidia AGX Orin SoC~\cite{malawade2021sage} has a Thermal Design Power (TDP) of 800W. These power demands can also increase the thermal demands on a vehicle's climate-control system. When combined, these demands can significantly reduce vehicle range~\cite{malawade2022ecofusion}.
This factor especially limits the mobility of electric vehicles due to their limited battery range and long recharge time. Further, it leads to higher vehicle usage costs and aggravates global energy and environmental issues. 
 
In general, the computational consumption for autonomous driving includes perception computing, decision-making computing, navigation, storage, etc~\cite{lin2018architectural}. Among them, perception computing is the most power-intensive component since it requires large neural networks to extract vehicle features (e.g., position and orientation) in the traffic environment. The mainstream perception architectures can be categorized into camera-only methods~\cite{liu2023sparsebev, ma2024vision, gan2024comprehensive} and multi-modality methods~\cite{liu2023bevfusion, liang2022bevfusion, xie2023sparsefusion}, where camera-only methods rely solely on image data from cameras, while multi-modality methods fusion the image and point cloud data from both cameras and LiDAR~\cite{gomez2022build}. Compared to camera-only methods, multi-modality methods have higher accuracy but incur significantly higher computational consumption. To alleviate this, researchers design model compression techniques to reduce the model size of multi-modality methods, including sparsification~\cite{xie2023sparsefusion}, quantization~\cite{zhang2023qd}, and distillation~\cite{wang2023distillbev}. However, they either maintain large computational consumption or exhibit significantly lower accuracy compared to the original multi-modality model in complex traffic scenarios, which may raise safety concerns of autonomous driving. Recently, a gate-based model~\cite{malawade2022ecofusion} is proposed to automatically discard or add parts of the network by gating techniques, thus altering the computational consumption and accuracy. Nonetheless, it struggles to balance computational consumption and accuracy, and training such a model is not trivial.


In summary, all the aforementioned methods reduce computational consumption by training a perception model with a reduced size. In this study, we aim to build a comprehensive and scalable perception optimization framework from the perspective of data management and tuning. 


(1) Firstly, instead of using a unified model for all traffic scenarios, we manage multiple trained models with different sizes tailored to different scenarios.
Secondly, we manage the framerate at which a perception model runs, which can directly reduce computational consumption without training additional models. In modern autonomous driving systems, the perception system typically operates at a framerate of 20-30 fps to extract environmental features~\cite{jebamikyous2022autonomous, caesar2020nuscenes}. We argue that such high framerates are not always necessary in all scenarios. For some simple scenarios (e.g., good weather condition and low traffic density), appropriately lowering the framerate does not compromise driving performance but can greatly reduce perception consumption. After frame reduction, we also consider using interpolation methods to fill in the features of skipped frames to keep the temporal granularity consistent.

\begin{figure}[!t]
    \centering
    \setlength{\abovecaptionskip}{0.2cm}
    \includegraphics[width=0.45\textwidth]{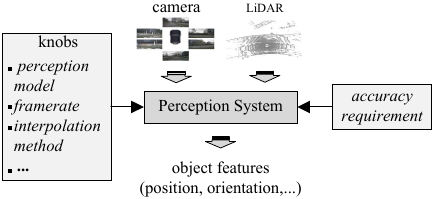}
    \caption{Knob Setting in Perception System}
    \label{intro}
    \vspace{-2mm}
\end{figure}

(2) As shown in Figure~\ref{intro}, we define the above adjustable parameters as tunable knobs (i.e., perception model, framerate, and interpolation method) in the perception system where each knob has a set of candidate values. We refer to a particular combination of knob values as a \textit{configuration} and there are $m^3$ configurations assuming each knob has $m$ candidate values. 
Different configurations have different levels of computational consumption and accuracy.
We set an accuracy requirement for feature extraction in the perception system and expect to find low-computation configurations while achieving the accuracy requirement. The optimal configuration of a traffic scenario is the one with the lowest computational consumption and over the accuracy.

However, implementing such a perception optimization framework within autonomous driving is non-trivial, which faces three main challenges: 

\emph{Challenge I: Diverse traffic scenarios.} The optimal configuration varies across different traffic scenarios. A low-computation configuration in simple scenarios is enough to achieve high accuracy, while in more complex traffic scenarios, a configuration with larger computation is required to maintain accuracy. 
Therefore, it is necessary to distinguish the perception difficulty of different traffic scenarios and separately search the corresponding optimal configurations.

\emph{Challenge II: Large configuration space.} The number of configurations $m^3$ grows rapidly with the number $m$ of candidate values for each knob. Although the computational consumption of a configuration can be estimated, obtaining accuracy requires running the configuration on hours or days of raw perception data. As a result, it is cost-prohibit to run all configurations in each scenario to find the optimal one.

\emph{Challenge III: Decision robustness.} Although there is an accuracy requirement during perception optimization, the extracted object features are inevitably perturbed. Therefore, the robustness of autonomous driving decisions needs to be enhanced to avoid some risky driving behaviors.





To tackle these challenges, we propose an energy-efficient autonomous driving framework, called \myfw, which includes an adaptive perception module and a robust decision module. 

(1) To address \emph{Challenge I}, a lightweight classification model is built to distinguish the perception difficulty of traffic scenarios, alongside an uncertainty value to enhance the model's reliability, where a scenario with a large uncertainty will be treated as the one with the highest difficulty. 
Ultimately, traffic scenarios are grouped into several difficulty levels, allowing the configuration optimization to focus on each level instead of on each single scenario.
(2) To address \emph{Challenge II}, we adopt Bayesian optimization instead of exhaustive search for knob tuning, which can find optimal/near-optimal configurations by evaluating only partial configurations. Further, we design a meta-learning strategy to transfer tuning knowledge across different types of traffic scenarios, further speeding up the tuning process. 
(3) To address \emph{Challenge III}, we model the decision-making task as a Markov decision process and design a regularization technique to prevent large policy updates, thus enhancing decision robustness.
To sum up, the adaptive perception module aims to reduce the energy consumption of the perception system, while the robust decision module aims to optimize driving decisions. In addition, since the robust decision module can lead to more conservative driving behaviors, we find that the energy consumption of the driving system also decreases to some degree. 


We summarize the main contributions as follows:

\noindent$\bullet$ We develop an energy-efficient autonomous driving framework, called \myfw, to adaptively reduce the energy consumption of the perception system while maintaining good driving performance.


\noindent$\bullet$ We tune the perception model, framerate, and interpolation method in the perception system and set them as knobs. Afterward, we design a meta-learning strategy within Bayesian optimization to explore promising configurations with low computational consumption and sufficient accuracy.

\noindent$\bullet$ We design a lightweight and reliable classification model to distinguish the perception difficulty of traffic scenarios.




\noindent$\bullet$ We design a regularization technique to enhance the decision robustness of autonomous driving.

\noindent$\bullet$ We conduct extensive experiments on real-world and synthetic datasets, evidencing the superiority of our proposals in both energy consumption and driving performance.



%% file: tex/related.tex
\section{Related Works}
\label{Related works}

Autonomous driving is recognized as a technology that could herald a major shift in transportation. Typically, there are two methods for building autonomous driving systems: modular and end-to-end. Modular methods leverage a perception module to extract traffic features from sensor data and then make driving decisions based on the perception results, while end-to-end methods attempt to directly map sensor data to driving decisions~\cite{coelho2022review}. Given that end-to-end methods struggle in complex driving scenarios and lack transparency and explainability~\cite{rosero2024integrating}, this study focuses on modular methods. In the following, we will introduce some related works about perception and decision-making and analyze their limitations. 

\noindent \textbf{Perception Module.}
Autonomous vehicles mainly rely on cameras and LiDAR for environmental perception~\cite{gomez2022build}. On the one hand, cameras produce RGB images that can be used to detect color and position features with relatively low computational consumption. However, their detection performance declines sharply under adverse weather and light conditions. On the other hand, while LiDAR demands a higher computational consumption and lacks color information, it can provide reliable depth information with point cloud data and is less affected by environmental factors. Based on them, there are two mainstream perception architectures: camera-only~\cite{liu2023sparsebev, ma2024vision, gan2024comprehensive}, and multi-modality~\cite{liu2023bevfusion, liang2022bevfusion, xie2023sparsefusion}, where camera-only methods rely solely on camera data, while multi-modality methods combine the sensor data from both cameras and LiDAR.

Although multi-modality methods can achieve higher perception accuracy, they incur significantly higher computational consumption compared to camera-only methods. Therefore, many studies use model compression techniques, including sparsification, quantization, and distillation, to lower computational consumption by reducing model sizes. 
However, they either continue to have a large model size or the accuracy is considerably inferior to that of the origin
multi-modality model. Recently, a gate-based model~\cite{malawade2022ecofusion} is proposed to adjust the model size of multi-modality methods by adaptively discarding or adding parts of the network. Nonetheless, it struggles to balance computational consumption and accuracy, and training such a model is not trivial.


To sum up, all these existing methods focus solely on training a model with a reduced size. In this study, we intend to optimize autonomous driving perception from the perspective of data management and tuning. Instead of using a unified model for all traffic scenarios, we manage multiple trained models with different sizes, and adjust the framerate at which a perception model runs and the interpolation method after frame reduction.



\noindent \textbf{Decision-making Module.}
After the perception module, the decision-making module is used to generate driving decisions based on the perception results.
In general, the decision-making methods of autonomous driving can be categorized into rule-based and data-driven. Rule-based methods~\cite{erdmann2015sumo, milanes2014modeling, xiao2017realistic} design a set of rule-matching algorithms and kinetic equations to calculate driving actions. However, these methods suffer from limited flexibility in handling complex scenarios with manually defined rules. With the development of artificial intelligence, many data-driven models use neural networks to generate driving actions. Researchers propose to build deep-learning models~\cite{chen2019deep, hussein2017imitation} to imitate the driving trajectories~\cite{chen2019real, deng2025exact, deng2024learning, ding2018ultraman,lian2023egl} of human drivers. Recently, the autonomous driving community has been increasingly embracing reinforcement learning methods~\cite{xia2024parameterized, xia2023smart, fu2022decision, liu2023impact,yu2024modus}, which are expected to outperform human drivers by continuous trial and error. 
The previous work~\cite{xia2024parameterized} proposes a reinforcement learning-based method with a parameterized action structure and a hybrid reward function to optimize the safety, efficiency, comfort, and impact on nearby traffic in autonomous driving.

Unfortunately, the aforementioned models are based on the assumption of a perfect perception of the traffic environment. In this study, we aim to construct a robust decision-making model when facing perturbed state features from the actual perception system.


%% file: tex/problem.tex
\section{Overview}
\label{Overview}
\subsection{Problem Statement}
In this study, we follow the standard perception and decision framework~\cite{xia2024parameterized} to construct the autonomous driving pipeline, where there is one autonomous vehicle $A$ and a set of conventional vehicles $\mathbb{C}$ traveling on multi-lane roads. For simplification, we do not consider pedestrians, traffic cones, or other obstacles. The \textbf{Input} is the image and point cloud data acquired from onboard sensors (i.e., six cameras and one LiDAR), where the perception module processes this data to extract environmental features. The \textbf{Output} is the steering angle and velocity control signals of the autonomous vehicle, which are calculated based on the extracted features.
Then, we describe some key metrics as follows:

\noindent \textbf{Energy Consumption.}
The autonomous vehicle in this study is electric-powered. In general, the perception system and driving system are the main components associated with electricity usage for autonomous driving. Firstly, the perception system involves running deep learning models to extract surrounding environmental features. 
To avoid interference from other processes and hardware factors (e.g., temperature), we measure its power consumption $Ene\text{-}P$ based on the total number of floating point operations ($\mathit{FLOPs}_{total}$) following the previous method~\cite{lin2018architectural, desislavov2021compute}, i.e., $Ene\text{-}P=\frac{\mathit{FLOPs}_{total}}{\eta}$, where $\eta=101.71 \mathit{GFLOPs}/J$ is the GPU efficiency of GeForce RTX 3090. 
Then, the driving system is used to drive the autonomous vehicle, where different driving policies exhibit varying levels of electricity usage. We calculate its power consumption $Ene\text{-}D$ based on the energy difference $E_D$ between two time steps~\cite{kurczveil2014implementation} and the energy loss $E_L$, i.e., $Ene\text{-}D=E_L+E_D(t_2)-E_D(t_1)$. $E_D(t)=\frac{m}{2}v_t^2+mgh_t+\frac{J_{int}}{2}v^2_t$, where $m$, $v_t$, $g$, $h_t$, and $J_{int}$ denotes vehicle mass, velocity, gravity acceleration, altitude, and equivalent moment of inertia respectively. $E_L$ is the energy loss caused by aerodynamic drag, rolling resistance, etc.


\begin{figure*}[t]
    \centering
    \setlength{\belowcaptionskip}{0.2cm}
    \includegraphics[width=0.92\textwidth]{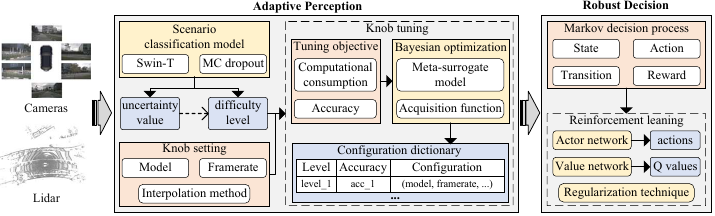}
    \caption{Framework Overview}
    \label{framework}
    \vspace{-2mm}
\end{figure*}

\noindent \textbf{Perception Accuracy.}
In addition to energy consumption, accuracy is another important metric of the perception module. Different feature extraction models have different levels of accuracy. In this study, we measure the accuracy by the nuscenes detection score (NDS), which is the most relevant accuracy metric for driving performance~\cite{schreier2023offline}. Specifically, $\footnotesize \mathrm{NDS}=\frac{1}{10} \allowbreak [5 \mathrm{mAP}$ $\allowbreak+\sum_{m \in\{\mathrm{mATE}, \mathrm{mASE}, \allowbreak \mathrm{mAOE}, \mathrm{mAVE}, \mathrm{mAAE}\}} \allowbreak (1-\min (1, m)) ]$, where $\mathrm{mAP}$, $\mathrm{mATE}$, $\mathrm{mASE}$, $\mathrm{mAOE}$, $\mathrm{mAVE}$, and $\mathrm{mAAE}$ denote mean average precision, translation error, scale error, orientation error, velocity error, attribute error, respectively.



\noindent \textbf{Driving Performance.}
We evaluate the driving performance from four factors: safety, efficiency, comfort, and impact, following the previous work~\cite{xia2024parameterized}. These factors are respectively measured by the autonomous vehicle's time-to-collision (TTC), speed, acceleration change, and deceleration of the vehicle behind it.


\noindent \textbf{Objective.}
Our objective is to adaptively reduce the energy consumption of the autonomous vehicle under a specific perception accuracy, while maintaining good driving performance.

\subsection{Framework Overview}
In this study, we propose an energy-efficient autonomous driving framework, called \myfw, which includes an adaptive perception module and a robust decision module. We show the framework overview in Figure~\ref{framework}.

\noindent \textbf{Adaptive Perception.}
This module aims to reduce the computational consumption of perception computing, which includes a classification model to output a perception difficulty level for each traffic scenario and a knob tuning component to explore promising configurations for each level. Specifically, the classification model is based solely on image data and applies a lightweight neural network Swin-T. Moreover, it also calculates an uncertainty value by Monte Carlo (MC) dropout. When the uncertainty value is greater than a threshold, the output difficulty level will be modified to the highest level to improve the model's reliability. In the knob tuning component, we first set the perception model, framerate, and interpolation method as knobs in the perception system. Then, Bayesian optimization is adopted for configuration search with two objectives: computational consumption and accuracy. To accelerate the tuning process, we further design a meta-surrogate model to transfer tuning knowledge across scenarios with different difficulty levels. Finally, the optimal configuration found at each level and accuracy requirement will be organized into a configuration dictionary. In the inference phase, the autonomous vehicle can directly choose a configuration from the dictionary based on a specified accuracy requirement and the difficulty level of the current traffic scenario without re-searching.

\noindent \textbf{Robust Decision.}
This module aims to make robust driving decisions when the autonomous vehicle uses the above perception module to perceive the surrounding traffic. Firstly, we model the decision-making as a Markov decision process (MDP) that includes four key elements: state, action, state transition, and reward. 
Then, we propose a reinforcement learning model to generate driving actions and design a regularization technique to prevent large policy updates, thus improving decision robustness. 



%% file: tex/adaptive_perception.tex
\section{Adaptive Perception}
\label{Adaptive Perception}
In this section, we first analyze the limitations of existing scenario classification models and introduce the development of our model. Then, we detail the knob setting in the perception module and the knob tuning model for exploring promising configurations in the configuration space.

\subsection{Limitation of Existing Scenario Classification Models}
The perception difficulty of each traffic scenario is influenced by various environmental factors, including weather conditions, lighting conditions, and traffic density~\cite{azfar2024deep}. In simple scenarios (e.g., sunny days), using only low-computation configurations can achieve a high perception accuracy, while in complex scenarios (e.g., rainy days), high-computation configurations are required to maintain accuracy. Since point cloud data cannot identify lighting conditions that can affect perception, researchers distinguish the perception difficulty of different traffic scenarios solely by the image data quality: higher quality means a lower difficulty level, while lower quality signifies a higher one. 
In general, the classification methods can be divided into traditional methods and deep learning methods. 
The traditional methods~\cite{gao2013universal, ghadiyaram2017perceptual} use nature scene statistics (NSS) to distinguish images, which can reflect the degree of image distortion based on some handcrafted metrics. With the development of neural networks, deep learning-based methods have drawn much attention for their superior performance. A recent study~\cite{ye2012unsupervised} proposes using neural networks to encode images and then adopting clustering methods to classify these encoding vectors. Due to the lack of interpretability in clustering rules, many studies~\cite{zhai2020perceptual, golestaneh2022no, yang2022maniqa} utilize deep learning-based models to directly predict the image quality. 
However, directly using these methods in our task usually leads to unsatisfactory results due to the following three limitations:

(1) They are general-purpose rather than specific to autonomous driving perception. Specifically, they annotate the image categories according to the degree of image distortion or subjective human scoring, which makes it difficult to accurately reflect the perception difficulty of different traffic scenarios. In contrast, a classification criterion directly related to the quality of subsequent perception results would be better for our scenario classification task.

(2) Although these methods can attain relatively high accuracy in scenario classification, their reliability is insufficient. When the autonomous vehicle encounters unseen traffic scenarios, these methods may underestimate the true difficulty level of the scenario. This can mislead our perception system to use low-computational configurations, which may result in a sharp decrease in accuracy and thus raise safety concerns.

\subsection{Development of Our Scenario Classification Model}
Based on the above analysis, we will build a task-specific and reliable difficulty classification model based on image data. To achieve this, we first build a dataset dedicated to the autonomous driving perception task, where the image data is automatically labeled based on the quality of perception results after running a unified perception model. Then, to avoid introducing a large computational overhead to the perception system, we use a lightweight model Swin-T~\cite{liu2021swin} as the main neural network to encode image data. Finally, to enhance the reliability of classification results, we calculate an uncertainty value using the Monte Carlo (MC)
dropout technique~\cite{jospin2022hands}, which is a simple and effective implementation of Bayesian neural networks. Next, we introduce the data generation process and model architecture as follows:


\noindent \textbf{Dataset Generation.}
Inspired by the previous work~\cite{kang2017noscope, kossmann2023extract}, we label the perception difficulty of images based on the detection results after running a unified object detection model on them. The images with better detection results are regarded as lower difficulty levels, and the ones with poorer results are regarded as higher levels.
To be detailed, we generate the dataset by the following three steps: 

(1) Collect image data from a real-world dataset Nuscenes-R~\cite{caesar2020nuscenes} and a high-fidelity synthetic dataset Nuscenes-S~\cite{acuna2021towards}, where the traffic scenarios have various weather conditions and vehicle density. Since the raw image data in each traffic scenario is based on a 360-degree bird's-eye view consisting of six images from different directions, we stitch these images together as one image instance. Afterward, we split all image instances into a training set and a test set in a ratio of 4 : 1. 

(2) Run a unified object detection model on raw image data in each traffic scenario and calculate a metric NDS~\cite{schreier2023offline} to evaluate detection performance.
Specifically, we choose SparseBEV~\cite{liu2023sparsebev} as the detection model, and NDS refers to the Nuscenes detection score that reflects perception accuracy. Generally, more complex scenarios (e.g., worse weather conditions) result in higher perception difficulty, which in turn leads to lower NDS values.

(3) Classify all image instances into $k$ difficulty levels based on their NDS values. We first sort all image instances by their NDS values in ascending order and then divide them into multiple equal groups for simplicity. Taking $k=4$ as an example, the difficulty levels 1, 2, 3, and 4 are allocated to four equal groups, and all image instances within a group are assigned the same difficulty level.

In the end, the scenario classification task in this study is defined as: Given an image instance $x$, predict the perception difficulty level $\hat{y}$ it belongs to, where $\hat{y} \in \{1,2,\dots,k\}$.

\begin{figure}[!t]
    \centering
    \setlength{\abovecaptionskip}{0.2cm}
    \includegraphics[width=0.45\textwidth]{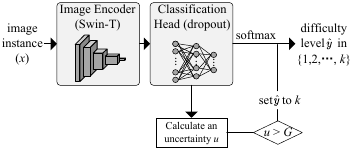}
    \caption{Model Architecture of Our Scenario Classification Model, which includes a Swin-T network to encode image data and a classification head with dropout layers to calculate a perception difficulty level $\hat{y}$ ($\hat{y} \in {1,2,\dots,k}$) of each image data $x$. In the inference phase, the classification head is run $T$ times to obtain an uncertainty value $u$. If $u$ is larger than a threshold $G$, the difficulty level will be set to $k$ (i.e., the highest difficulty level).}
    \label{classify}
\end{figure}

\noindent \textbf{Network Architecture.}
We introduce the network architecture of our scenario classification model as shown in Figure~\ref{classify}, and present the calculation formulas in the following.  

Firstly, we design an image encoder that uses the Swin-T network~\cite{liu2021swin} to encode the image sample $x$, as follows:

\begin{equation}
\footnotesize
    h = \textit{Swin-T} (x, w_1),
\end{equation}
where $h$ denotes the encoding vector of $x$ and $w_1$ denotes the learnable network parameters. In the experiment Section~\ref{Evaluation of Adaptive Perception Module}, we find that this lightweight model Swin-T achieves almost the same accuracy as other heavyweight networks.

Then, we design a classification head based on Multi-layer Perceptron (MLP) to calculate the probability that the image data $x$ belongs to each difficulty level, as follows:

\begin{equation}
\footnotesize
P = \mathit{softmax} (\mathit{MLP} (h, w_2)),
\label{probability vector}
\end{equation} 
where $w_2$ denotes the learnable network parameters, $P=[p_1, p_2, \dots, p_k]$ is a probability vector that respectively indicates the probabilities of $k$ difficulty levels, and the level with the highest probability is regarded as the predicted level of $x$. We use three linear layers in this MLP. Moreover, we use the MC dropout technique~\cite{gal2016dropout} in the MLP to avoid model overfitting in the training phase and obtain uncertainty in the inference phase.

In the training phase, since our scenario classification is a multi-class classification task, we use the cross-entropy loss~\cite{mao2023cross} as our loss function $\mathcal{L}_1$, as follows:
\begin{equation}
    \begin{gathered}
    \footnotesize
\mathcal{L}_1=-log(p_y),
\end{gathered}
\end{equation}
where $y$ denotes the true difficulty level of the image data $x$, and $p_y$ is the probability assigned to the true level in the probability vector $P$. This loss encourages the classification model to maximize the probability of the true level.

In the inference phase, the classification head performs $T$ forward pass after encoding an image $x$ in the image encoder, where each forward pass randomly deactivates a specific fraction of neurons. Afterward, we calculate the difficulty level $\hat{y}$ of $x$ and an uncertainty value $u$, as follows:

\begin{equation}
    \begin{gathered}
    \footnotesize
    \hat{y}=\mathit{argmax}(\frac{1}{T}\sum_{i=1}^{T}P^i), \\
    u = \frac{1}{T}\sum_{i=1}^{T}(p^{i}_{\hat{y}}-\frac{1}{T}\sum_{i=1}^{T}p^{i}_{\hat{y}})^2,
\end{gathered}
\end{equation}
where $P^i$ denotes the probability vector in the $i-$th forward pass based on Equation~\ref{probability vector}, $\mathit{argmax}$ is used to find the difficulty level with the highest average probability as the final classification result $\hat{y}$, and $p_{\hat{y}}^i$ denotes the probability of $\hat{y}$ at the $i-$th forward pass.
If the uncertainty value $u$ is larger than a threshold $G$, we will set the difficulty level $\hat{y}$ to $k$ (i.e., the highest difficulty level). In the later knob tuning section, the traffic scenarios with a difficulty level of $k$ will use the highest-computation configuration.
Therefore, this operation helps prevent a sharp decline in the perception accuracy caused by incorrect scenario classification results. Also, the additional computation it brings is minimal because we only run the classification head $T$ times, excluding the image encoder.


\subsection{Knob Setting for Adaptive Perception}
\label{Knob Setting for Adaptive Perception}
After the above section, all traffic scenarios are classified into $k$ difficulty levels. For each type, we respectively set three knobs (i.e., perception model, framerate, and interpolation method after frame reduction) that can affect both computational consumption and perception accuracy. To be detailed, we describe these knobs as follows:

(1) $knob_1$: Perception models with different sizes.
This study adopts a series of BEV-based models to extract the surrounding vehicle features (e.g., position and orientation) of the autonomous vehicle. 
Specifically, we set the candidate set of $knob_1$ as $\{\mathit{SparseBev}, \allowbreak \mathit{SparseFusion}, \mathit{BevFusion}, \mathit{BevFusion}\mathord{-}e\}$, where SparseBEV~\cite{liu2023sparsebev} is a camera-only method, SparseFusion~\cite{xie2023sparsefusion} and BevFusion~\cite{liu2023bevfusion} are multi-modality methods, and BevFusion-e is the ensemble version of BevFusion, respectively. We implement them based on their open-source code repositories. From left to right, these models have higher computational consumption but higher accuracy.




(2) $knob_2$: Framerate. The onboard cameras and LiDAR in this study capture image data and point cloud data at 20 frames per second. Based on it, we express the frame rate as the number of frames to skip each time before the next run of a perception model. Then, the candidate set for this knob is defined as $\{0,1,2,3,4,5,6,7,8,9\}$ frames. A larger value results in larger resource savings but larger accuracy degradation.

(3) $knob_3$: Interpolation methods after frame reduction. If the perception framerate is reduced, this study considers using a interpolation method to fill in the vehicle features in skipped frames. The candidate set of $knob_3$ is set as $\{linear, prediction\}$. The linear method assumes that the surrounding vehicle maintains the driving actions of the last time frame~\cite{wang2023we}. 
On the contrary, the prediction method fills in the missed vehicle features with a trajectory prediction model, following the previous work~\cite{varadarajan2022multipath++}. The prediction method has higher computational consumption but higher accuracy for the filled features compared to the linear method. 

Finally, the number $n$ of configurations can be calculated as the product of the number of candidate values for these three knobs at $k$ different scenario difficulty levels, i.e., $n=4\times10\times2\times k=80k$. Although the computational consumption of a configuration can be estimated, its accuracy is unpredictable and needs to be obtained by running it on raw perception data (i.e., images and/or point cloud data), which demands much computational consumption~\footnote{We run these configurations on a 5.5-hour dataset, and each configuration takes an average of 14 hours on a GeForce RTX 3090.}. Therefore, it is cost-prohibit to run all configurations to get the optimal configuration in each scenario difficulty level. 


\subsection{Knob tuning for Finding Promising Configurations}
\label{Knob tuning for Finding Promising Configurations}
Based on the above, we adopt knob tuning to find optimal/near-optimal configurations by evaluating partial configurations rather than all configurations. Also, there are $k$ \textit{tuning instances} corresponding to traffic scenarios of $k$ perception difficulty levels.
This study adopts multi-objective Bayesian optimization to perform knob tuning, given its superior performance in other tuning tasks~\cite{zhao2023automatic}. To further speed up the tuning process, we design a meta-surrogate model to transfer tuning knowledge between different difficulty levels.


We introduce the main components of Bayesian optimization in each tuning instance as follows:

(1) Objective Function $f(\theta)$. We use $\theta_j$ to represent a configuration and $f^i(\theta)$ to represent the objective function in a tuning instance $i$. In this study, the objective function consists of two objectives, i.e., computational consumption $f^i(\theta).com$ and accuracy $f^i(\theta).acc$. To align both objective values with the minimization goal of Bayesian optimization, $f^i(\theta).acc$ is the negated value of raw accuracy value.


(2) Surrogate Model $\mathcal{M}$. We represent the surrogate model in a tuning instance $i$ as $\mathcal{M}^i=(\theta, f^i(\theta), \sigma^2)$, which can approximate the objective function value $f^i(\theta)$ of each configuration and provide an uncertainty estimation $\sigma^2$. It can be continuously updated according to the objective function values of actual evaluated configurations. Since the knob value is discrete, we adopt SMAC (Sequential Model-based Algorithm Configuration)~\cite{hutter2011sequential} as the probability surrogate model. 

(3) Acquisition function $\alpha$. The acquisition function $\alpha$ is used to decide which configuration to explore based on the surrogate model. To adapt to multiple objectives, we use EHVI (Expected Hypervolume Improvement)~\cite{li2021openbox} as the acquisition function.


At each tuning iteration, we first obtain a configuration using the acquisition function $\alpha^i$. Then, we run the configuration to get its objective function values, based on which the surrogate model is updated. As more tuning iterations are done, the surrogate model becomes more accurate, and the acquisition function directs the search toward the optimal solution. 

\noindent \textbf{Meta-surrogate Model.}
Based on the above, there are $k$ independent tuning instances for traffic scenarios with different perception difficulty levels. Each tuning instance needs to explore and run explored configurations separately and update its surrogate model from scratch. In this study, we design a meta-surrogate model to reduce the number of tuning iterations by warm-starting a new tuning instance with the previous tuned instances. However, directly transferring the surrogate model from other tuning instances is not feasible, as their output accuracy values differ in scale.
That is, the magnitude of accuracy values for the same configuration diverges sharply across tuning instances of scenarios with different perception difficulty levels.
To address this, we adopt a simple strategy that normalizes the accuracy value within each tuning instance. As a result, the output accuracy of each surrogate model is a relative value rather than an absolute one. The relative relationships in the objective value between configurations can still provide valuable insights in different tuning instances~\cite{zhang2021restune}. Finally, we will detail the main steps of this transfer strategy.

(1) Normalize objective values. To normalize the objective function in a tuning instance $i$, we first run the lowest-computation configuration $\theta_l$ and the highest-computation configuration $\theta_h$ to get their objective values. Then, we use min-max normalization to normalize the objective function values as $\overline{f}^i(\theta_j).com$ and $\overline{f}^i(\theta_j).acc$.



(2) Construct meta-surrogate model.
We use $\mathcal{M}^i$ ($i=1,2,\dots, k$) to represent the surrogate models in $k$ tuning instances and execute them sequentially. The first tuning instance still starts tuning from scratch. But for subsequent tuning tasks, we build a meta-surrogate model $\mathcal{M}^{meta}$ using the previously tuned instances. For example, assuming we will execute the $i$-th ($i>1$) tuning instances, we initialize the objective function $\overline{f}^{meta}(\theta)$ as follows.

\begin{equation}
    \begin{gathered}
    \footnotesize
     \overline{f}^{meta}(\theta)= \frac{\sum_{b=1}^{i-1}\overline{f}^{b}(\theta)}{i-1}, i > 1,\\
\end{gathered}
\end{equation}
where $\overline{f}^{meta}(\theta)$ is the average of the objective functions of all previously tuned tasks, and we use it to initialize the surrogate model $\mathcal{M}^i$. In this way, $\mathcal{M}^i$ can be represented as $\mathcal{M}^i=\mathcal{M}^{meta}=(\theta, \overline{f}^{meta}(\theta), \sigma^2)$. Note that the uncertainty estimation $\sigma^2$ is initialized with the default method~\cite{hutter2011sequential} to guide the exploration in the configuration space.

\begin{algorithm}[!t]
\small
    \caption{Knob Tuning with A Meta-surrogate Model}
    \label{algorithm1}
    \begin{algorithmic}[1]
        \Require{$k$ tuning instances corresponding to traffic scenarios of $k$ perception difficulty levels, the initial surrogate model $\mathcal{M}^i\ (i=1,2,\dots,k)$, the acquisition function $\alpha$, the maximum number of configurations $N$, the accuracy requirements $\mathcal{A}_i\ (i=1,2,3,4,5)$, the configuration dictionary $\mathbb{D}$ }
        \For{$i=1$ to $k$}
        \If{$i>1$}
        \State \textit{// construct the meta-surrogate model} 
        \State $\overline{f}^{meta}(\theta)= \frac{\sum_{b=1}^{i-1}\overline{f}^{b}(\theta)}{i-1}$
        \State $\mathcal{M}^i=\mathcal{M}^{meta}=(\theta, \overline{f}^{meta}(\theta), \sigma^2)$
        \EndIf
        \State $j=1$
        \State run the highest-computation and lowest-computational 
        \State configurations $\theta_l$ and $\theta_h$
        \While{$j<N$ \&\& no convergence}
        \State \textit{// using the acquisition function $\alpha$}
        \State get a configuration $\theta_j$ 
        \State \textit{// run $\theta_j$}
        \State get the objective values $f^i(\theta_j).com$ and $f^i(\theta_j).acc$ 
        \State update the configuration dictionary $\mathbb{D}$
        \State normalize the objective values as $\overline{f}^i(\theta_j).com$ 
        \State and $\overline{f}^i(\theta_j).acc$
        \State update the surrogate model $\mathcal{M}^i$,
        \State increment $j$,
        \EndWhile
        \EndFor
    \end{algorithmic}
\end{algorithm}

We present the entire tuning process in Algorithm~\ref{algorithm1}. At first, we define five accuracy requirements, $\mathcal{A}_i \ (i=1,2,3,4,5,6,7,8)$, corresponding to NDS values of 0.69, 0.70, 0.71, 0.72, 0.73, 0.74, 0.75, and 0.76~\footnote{The most expensive model BevFusion-e~\cite{liu2023bevfusion} in our experiment can only achieve an NDS of 0.761. After the experiment, we chose 0.74 as the final accuracy requirement, which achieves a good balance between energy consumption and driving performance.}, respectively. 
Then, the pseudocode in lines $2-5$ shows the meta-surrogate model construction at the beginning of a tuning instance. The pseudocode in lines $6-18$ shows the tuning process, where the acquisition function is used to select the next configuration to explore, and the surrogate model is updated with normalized objective values. During tuning, we maintain a configuration dictionary $\mathbb{D}$ to save the current lowest-computation configurations under different accuracy requirements in different tuning instances. If all configurations fail to reach a certain accuracy in a tuning instance, its configuration in the dictionary will be set to the highest-computation one. Finally, we will terminate a tuning instance when it converges or the maximum number of configurations $N = 80$ is reached. 




%% file: tex/robust_decision.tex
\section{Robust Decision}
\label{Robust Decision}
This section is used to make driving decisions based on perception results from the above perception module. Firstly, we model the decision-making problem under Markov decision process (MDP).
Then, we design a regularization technique to enhance the decision-making robustness when facing perturbed state features. 

\subsection{Markov Decision Process}
Markov decision process (MDP) can be defined as: $\mathcal{M} = \langle{\mathcal{S}, \mathcal{A}}, \mathcal{P}, \mathcal{R}\rangle$, which includes four key elements: state $\mathcal{S}$, action $\mathcal{A}$, state transition probability $\mathcal{P}$ and reward $\mathcal{R}$. Specifically, the state $s^t$ denotes the extracted environmental features at time frame $t$, the action $a^t$ denotes the driving decision, the state transition is implemented via environment rendering in the simulator Carla~\cite{Dosovitskiy17}, and the reward $r^t$ is used to rate the action, where we adopt a hybrid reward function including four reward factors: safety, efficiency, comfort, and impact. Following the previous work~\cite{liu2023impact}, the safety factor is measured by time-to-collision (TTC) of the autonomous vehicle, which denotes the time span left before a collision. The efficiency factor and the comfort factor are measured by the velocity and acceleration change rate of the autonomous vehicle. The impact factor is measured by the passive deceleration of the rear traffic flow caused by the autonomous vehicle. 

\begin{figure}[!t]
    \centering
    \setlength{\abovecaptionskip}{0.2cm}
    \includegraphics[width=0.45\textwidth]{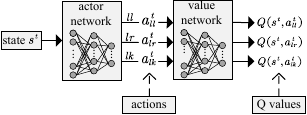}
    \caption{Model Architecture of Our Reinforcement Learning Model. The actor network outputs three action values corresponding to three lane-changing decisions, and the value network outputs the Q values of the three actions. The action with the highest Q value is the final action performed by the autonomous vehicle.}
    \label{reinforce}
\end{figure}

\subsection{Reinforcement Learning with Regularization Technique}
Figure~\ref{reinforce} shows the architecture of our reinforcement learning model, which includes an action network and a value network. Specifically, the actor network incorporates a parameterized action schema~\cite{xiong2018parametrized}, outputting three action values (i.e., $a_{ll}^t$, $a_{lr}^t$, and $a_{lk}^t$~\footnote{Each action includes a steering angle value and a velocity change value.}) at each time step that correspond to three lane-changing decisions (i.e., left lane-changing $ll$, right lane-changing $lr$, and lane-keeping $lk$). The value network outputs the Q values of the three actions, and the action with the highest Q value is the final action performed by the autonomous vehicle.

To solve the parameterized action schema, researchers propose many reinforcement learning methods~\cite{masson2016reinforcement, hausknecht2015deep,xiong2018parametrized, xia2024parameterized, zeng2023target} to learn an action policy. Specifically, the P-DDPG method~\cite{hausknecht2015deep} collapses the parameterized action space into a continuous one. However, it does not account for which action is associated with which lane-changing decision, leading to suboptimal action policies. Recently, the P-DQN method~\cite{xiong2018parametrized} is proposed to directly generate multiple actions, and each action is associated with a decision. Benefiting from this, P-DQN can learn a better action policy under the parameterized action structure. To improve the robustness when facing perturbed state features, the RP-DQN method~\cite{xia2024parameterized} adds a regularization term into the actor network to stabilize the update process of the action network. However, this method does not consider the extrapolation error that can lead to the misestimation problem of Q values of state-action pairs~\cite{fujimoto2021minimalist}. Specifically, the Q value in reinforcement learning is used to estimate the expected reward when performing an action based on certain state features. Since the values of state features and action are continuous in this work, the number of state-action pairs is infinite. During the learning process, the reinforcement learning agent inevitably encounters some state-action pairs that significantly differ from those it has previously seen. In such cases, the agent tends to estimate their Q-values incorrectly. This in turn affects policy improvement, where the agent learns to prefer poor actions, leading to potential safety risks in autonomous driving. Therefore, this study proposes a regularization technique to restrain the learning process of Q values, making the exploration of state-action pairs stay close to those previously seen.
Next, we will introduce the loss functions with a regularization term in the actor and value networks.


We represent an explored experience as $e^t=(s^t, a^t_{h_1}, r^t, s^{t+1}, a^{t+1}_{h_2})$ that denote the current state, action, reward, the next state and action, respectively. $h_1$ and $h_2$ denote the selected lane-changing decisions of $a^t_{h_1}$ and $a^t_{h_2}$, respectively. In the training process, the value network aims to minimize the gap between the estimated Q value $Q(s^t, a^t_h)$ and the target Q value $y$, and the loss function $\mathcal{L}_c$ is calculated as follows:

\begin{equation}
\label{equation-critic}
    \begin{gathered}
    \footnotesize
    y=r^t+\gamma (Q(s^{t+1}, \pi(s^{t+1}))-\beta_1(\pi(s^{t+1})-a_{h_2}^{t+1})^2),
    \\
    \mathcal{L}_c = \mathbb{E}_{D}(y-Q(s^t, a_{h_1}^t))^2,
    \end{gathered}
\end{equation}
where $\pi$ denotes the action policy learned in the actor network, $\gamma$ denotes the discount factor, and $D$ denotes a batch of experiences. Importantly, $(\pi(s^{t+1})-a_{h_2}^{t+1})$ is the regularization term that penalizes the agent based on how much the current output action $\pi(s^{t+1})$ deviates from the previously explored action $a_{h_2}^{t+1}$ in a state $s^{t+1}$. It is inspired by the behavior cloning technique~\cite{tarasov2024revisiting} that aims to make $\pi(s^{t+1})$ stay close to $a_{h_2}^{t+1}$, thus alleviating the misestimation problem of Q values. $\beta_1$ is the coefficient that determines the regularization scale. In addition to the value network, the actor network aims to learn an action policy to output the action with the largest Q value. We use the loss function in the previous work~\cite{xia2024parameterized} to learn the action policy.



In the end, we find that the regularization not only stabilizes the learning process in the face of perturbed state features, but also makes the final decision performance more conservative. This will lead to a slight degradation in efficiency, but the safety, comfort, and energy consumption of the driving system will be optimized.

%% file: tex/experiment.tex
\section{Experiments}
\label{Experiments}
\subsection{Experimental Settings}
\noindent \textbf{Dataset.}
We simulate the entire autonomous driving pipeline on the Carla simulator~\footnote{https://carla.org/}, which is a widely used project focused on creating a publicly available virtual environment for autonomous driving~\cite{Dosovitskiy17}. It supports almost all sensors (e.g., camera and LiDAR) with the goal of flexibility and realism in high-fidelity simulations. In the adaptive perception module, we use both a real-world dataset Nuscenes-R~\cite{caesar2020nuscenes} and a synthetic dataset Nuscenes-S~\cite{acuna2021towards}. Specifically, Nuscenes-R has 5.5-hour perception data where the framerate of annotated scenarios is only 2 fps. Nuscenes-S is generated with Carla, simulating the sensor and scenario setting in Nuscenes-R but with annotated scenarios at 20 fps. We use both datasets to train and test our scenario difficulty classification model but only use Nuscenes-S for knob tuning since the framerate of annotated scenes in Nuscenes-R is too low. For our decision-making model, we use the simulated dataset REAL in the previous work~\cite{xia2024parameterized}, which contains 55 different driving routes with various weather conditions and traffic density. Finally, we split all the above datasets into training sets and test sets with a splitting ratio of 4 : 1.

\noindent \textbf{Implementation Details.}
In the adaptive perception module, we set the number $k$ of perception difficulty levels in our scenario classification model to 4. For the classification head, we use three linear layers in the multi-layer perceptron, and their dimensions are set as 256, 128, and 64, respectively. To calculate uncertainty, the dropout rate is set as 0.5, the classification head performs 10 forward passes, and the threshold $G$ of uncertainty values is set as 0.02. To the end, we train the classification model using the Adam optimizer~\cite{kingma2014adam} with a learning rate of 0.01 and a batch size of 128. In the knob tuning section, the accuracy requirements $\mathcal{A}_i\ (i=1,2,3,4,5,6,7,8)$ are set to 0.69, 0.70, 0.71, 0.72, 0.73, 0.74, 0.75, and 0.76, respectively.
In the robust decision-making module, the actor network and value network use a
soft update mechanism and an updated ratio of 0.01 similar to DDPG~\cite{lillicrap2015continuous}, and the dimensions of the linear layers in the multi-layer perceptron are set to 64.
The reward discount factor $\gamma$ in the Bellman function is 0.9, and the coefficient $\beta_1$ in the regularization term is set to 0.4. In the end, we train the reinforcement learning model using the Adam optimizer~\cite{kingma2014adam} with a learning rate of 0.001 and a batch size of 128. Our experiments are conducted on NVIDIA RTX 3090 GPUs.

\noindent \textbf{Baselines.}
We compared our framework with five methods, including one standard perception-and-decision autonomous driving method and four model compression methods.

(1) Auto~\cite{xia2024parameterized}. 
A standard perception-and-decision framework for autonomous driving, which perceives the traffic environment with BevFusion~\cite{liu2023bevfusion} and makes driving decisions by a reinforcement learning model RBP-DQN.

(2) Sparse~\cite{liu2023sparsebev}.
It reduces computing consumption by using sparse representations that only focus on the relevant and non-empty areas of BEV representations.

(3) Quantize~\cite{zhang2023qd}.
It reduces computing consumption by lowering the precision of floating-point numbers from single-precision to half-precision.

(4) Distill~\cite{wang2023distillbev}.
It reduces computing consumption by distilling knowledge from a large multi-modality model to a smaller model.

(5) EcoFusion~\cite{malawade2022ecofusion}.
It can vary the model size of multi-modality methods by discarding or adding
parts of the network, thus altering computational consumption and accuracy.

However, Sparse, Quantize, Distill, and EcoFusion only focus on perception optimization but do not consider the impact of reduced perception computing on subsequent autonomous driving decision performance. Therefore, we plug them into the autonomous driving pipeline of Auto to evaluate their driving performance.

\begin{figure*}[!t]
\centering
\begin{minipage}[b]{0.23\linewidth}
\centering
\setlength{\abovecaptionskip}{0.1cm}
\includegraphics[width=\linewidth]{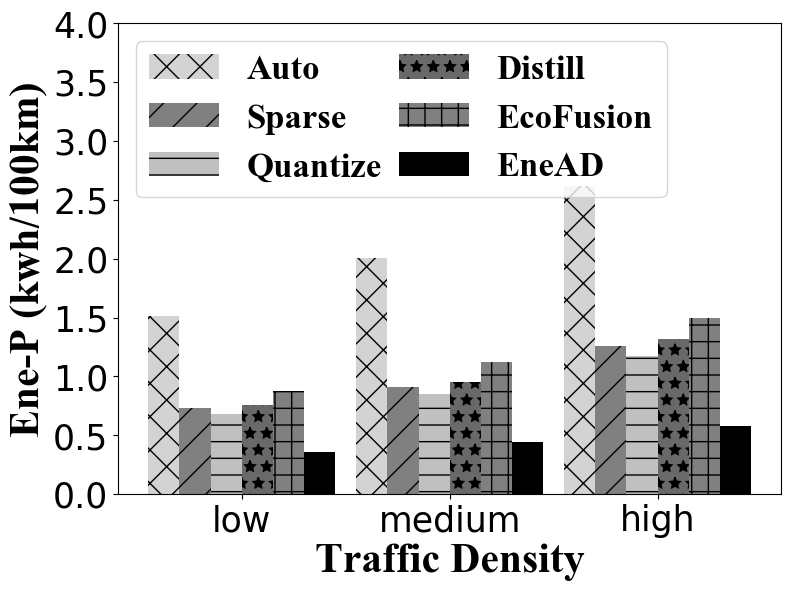}
\captionof{figure}{Evaluation of Perception Consumption}  
\label{figure: Evaluation of Perception Consumption}
\end{minipage}
\begin{minipage}[b]{0.23\linewidth}
\centering
\setlength{\abovecaptionskip}{0.1cm}
\includegraphics[width=\linewidth]{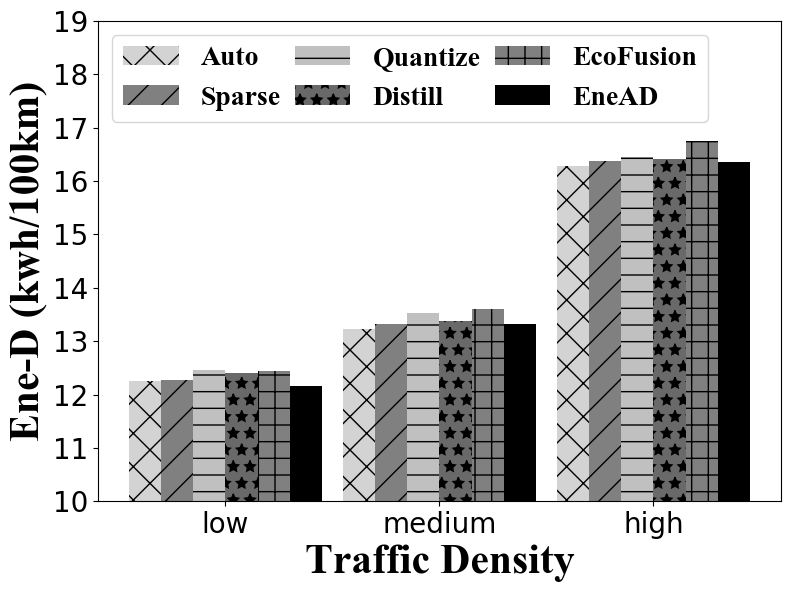}
\captionof{figure}{Evaluation of Driving Consumption}
\label{figure: Evaluation of Driving Consumption}
\end{minipage}
\begin{minipage}[b]{0.23\linewidth}
\centering
\setlength{\abovecaptionskip}{0.1cm}
\includegraphics[width=\linewidth]{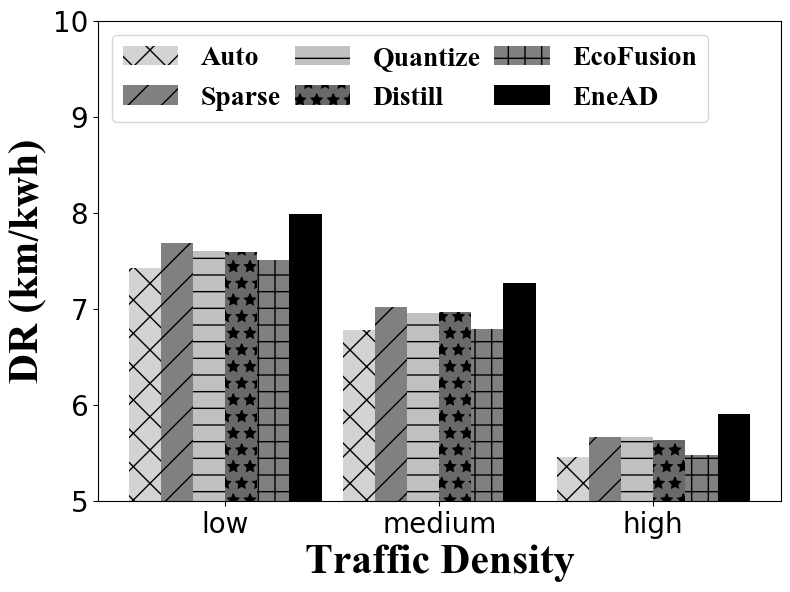}
\captionof{figure}{Evaluation of Driving Range}
\label{figure: Evaluation of Driving Range}
\end{minipage}
\begin{minipage}[b]{0.23\linewidth}
\small
    \centering
    \begin{tabular}{|p{1.8cm}<{\centering}|p{1.4cm}<{\centering}|}
         \hline
         \tabincell{c}{Metric} & \tabincell{c}{Percentage} \\
         \hline \hline
         \tabincell{c}{Overestimated} & 35.6 \\\hline
         \tabincell{c}{Underestimated} & 52.1\\\hline
         \tabincell{c}{Correct} & 12.3\\\hline
         
         \end{tabular}
\captionof{table}{Uncertainty Analysis of Scenario Difficulty Classification Model}
    \label{table of uncertainty of classification model}
\end{minipage}
\vspace{-2mm}
\end{figure*}

\begin{figure*}[!t]
\centering
\begin{minipage}[b]{0.23\linewidth}
\centering
\setlength{\abovecaptionskip}{0.1cm}
\includegraphics[width=\linewidth]{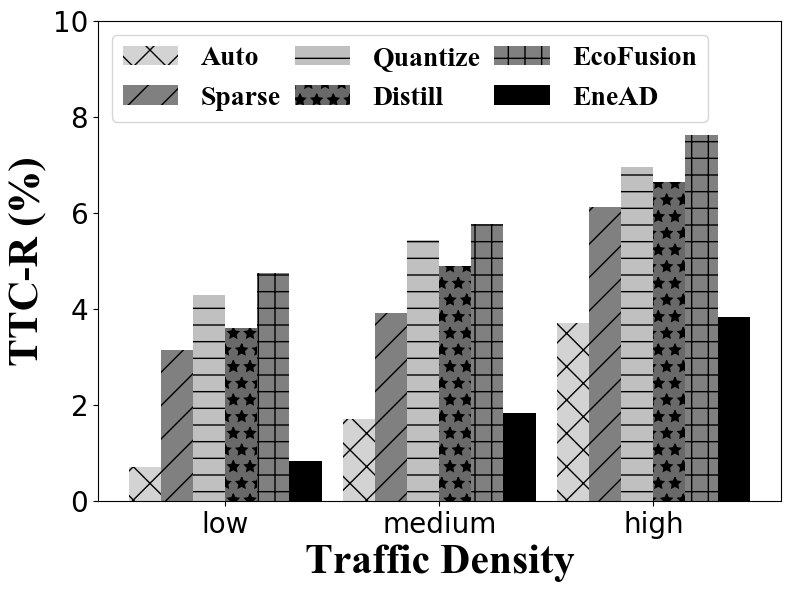}
\captionof{figure}{Safety Evaluation}  
\label{figure: Safety Evaluation}
\end{minipage}
\begin{minipage}[b]{0.23\linewidth}
\centering
\setlength{\abovecaptionskip}{0.1cm}
\includegraphics[width=\linewidth]{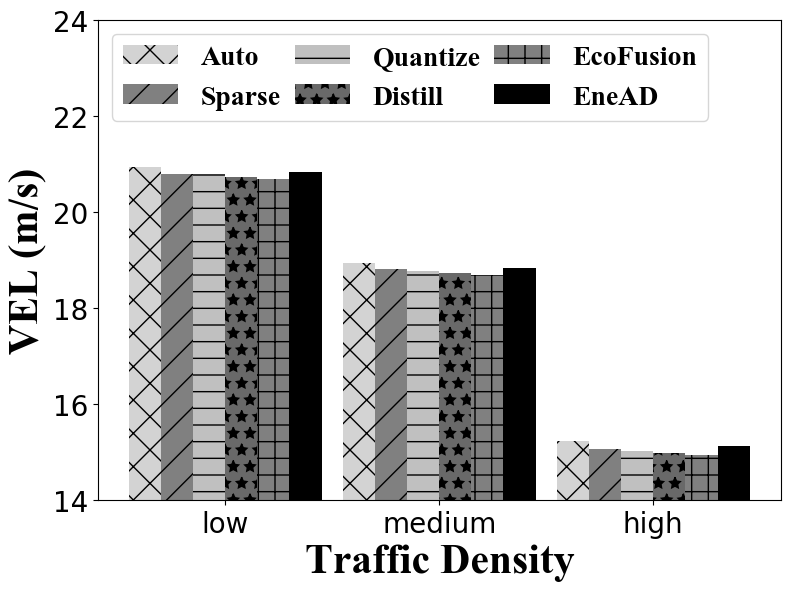}
\captionof{figure}{Efficiency Evaluation}
\label{figure: Efficiency Evaluation}
\end{minipage}
\begin{minipage}[b]{0.23\linewidth}
\centering
\setlength{\abovecaptionskip}{0.1cm}
\includegraphics[width=\linewidth]{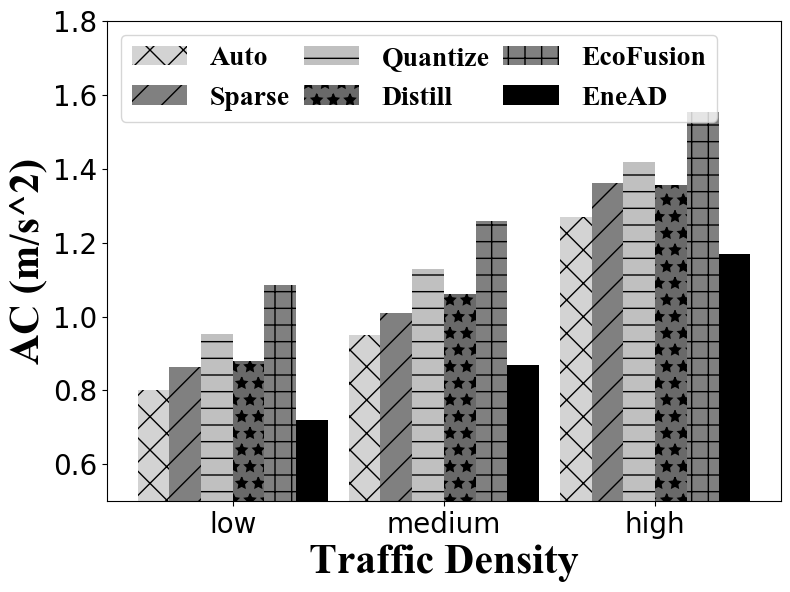}
\captionof{figure}{Comfort Evaluation}
\label{figure: Comfort Evaluation}
\end{minipage}
\begin{minipage}[b]{0.23\linewidth}
\centering
\setlength{\abovecaptionskip}{0.1cm}
\includegraphics[width=\linewidth]{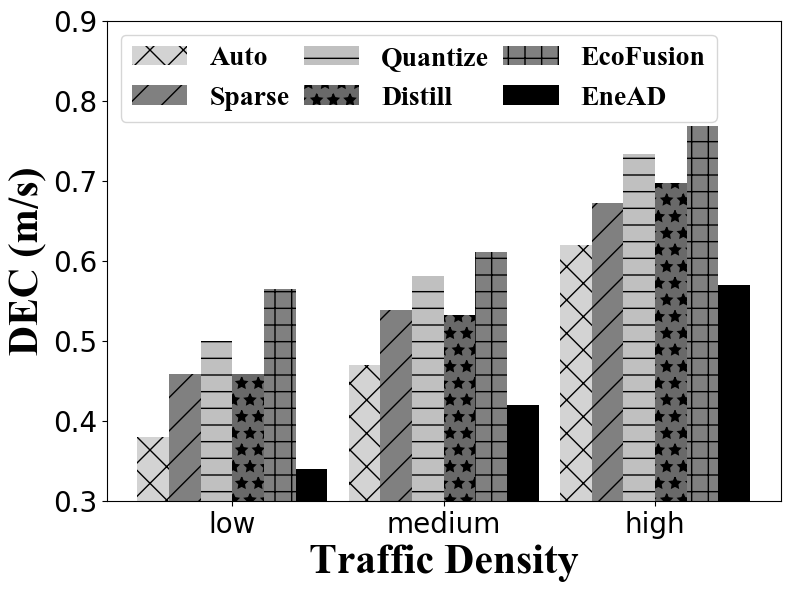}
\captionof{figure}{Impact Evaluation}
\label{figure: Impact Evaluation}
\end{minipage}
\vspace{-2mm}
\end{figure*}

\subsection{End-to-end Evaluation}
In this section, we study the end-to-end performance~\footnote{The result is based on the accuracy requirement of 0.74.} of our framework \myfw\ by comparing it against the baselines from two aspects: energy consumption and driving performance. 
Firstly, we design three metrics to evaluate the energy consumption of the autonomous vehicle, as follows:

(1) Average Energy consumption of perception system (Ene-P). We record the power consumption per 100 kilometers brought by the perception system of the autonomous vehicle. To avoid interference from other processes and hardware factors, we measure it based on the number of floating point operations (FLOPs) following the
previous method~\cite{lin2018architectural, desislavov2021compute}. We also use equivalent computational units for method Quantize, where one FP16 operation is counted as 0.5 FP32 operations 

(2) Average Energy consumption of driving system (Ene-D). We record the power consumption per 100 kilometers (km) brought by the driving system of the autonomous vehicle. It is influenced by many factors, such as vehicle speed and acceleration, and we obtain it by the SUMO platform following the previous method~\cite{kurczveil2014implementation}.

(3) Average Driving range (DR). We record the driving range (km) of the autonomous vehicle per kwh, considering only the energy consumption of the perception system and driving system.

Then, we design four metrics, following the previous work~\cite{xia2022rise,liu2023impact}, to evaluate the driving performance (i.e., safety, efficiency, comfort, and impact) of the autonomous vehicle, as follows:

(1)
Percentage of time-to-collision values with potential collision risks (TTC-R). We record the time-to-collision values of the autonomous vehicle and count the percentage of TTCs less than 4 seconds that are regarded as having potential collision risk following the previous work~\cite{xia2022rise}. A smaller TTC-R indicates that the autonomous vehicle is safer.

(2)
Average velocity (VEL).
We record the average velocity of the autonomous vehicle. A larger VEL indicates higher traffic efficiency.

(3)
Average acceleration change (AC).
We record the average acceleration change of the autonomous vehicle. A smaller AC indicates a more comfortable driving experience.

(4)
Average deceleration (DEC). After the autonomous vehicle performs abrupt braking or lane-changing behaviors, we record the deceleration of the rear vehicle. 
A smaller DEC means that the autonomous vehicle has less impact on the rear vehicle.

To demonstrate the effectiveness of our framework, we present these metrics under different traffic densities (i.e., low: $<10veh/km/ln$, medium: $10-30veh/km/ln$, and high: $>30veh/km/ln$).

Firstly, we report the metrics of energy consumption in Figure~\ref{figure: Evaluation of Perception Consumption},~\ref{figure: Evaluation of Driving Consumption}, and~\ref{figure: Evaluation of Driving Range}. For power consumption of the perception system in Figure~\ref{figure: Evaluation of Perception Consumption}, we can see that our framework \myfw\ achieves the smallest Ene-P, demonstrating that \myfw\ can achieve the lowest perception consumption. Among these baselines, the Auto method has the largest Ene-P since it does not consider reducing the perception consumption. In addition, although the model compression methods Sparse, Quantize, Distill, and EcoFusion can achieve a smaller Ene-P than Auto, they also have larger Ene-P values than our framework since they do not consider adjusting the framerate at which perception models run. 
Afterward, in Figure~\ref{figure: Evaluation of Driving Consumption} we can see that our framework \myfw\ has the smallest Ene-D, which demonstrates that \myfw\ can also lower the power consumption of the driving system. This is because our robust decision module can generate relatively conservative driving behaviors with smoother velocity changes. 
The poor performance of Sparse, Quantize, Distill, and EcoFusion is attributed to their inferior perception accuracies in challenging traffic scenarios, leading to more sudden speed changes that can increase power consumption. 
Finally, we show the driving range metric DR in Figure~\ref{figure: Evaluation of Driving Range}. Since \myfw\ has the lowest power consumption of both the perception and driving system, it can achieve the largest driving range per kwh. 
To sum up, our framework \myfw\ can achieve a $1.9\times-3.5\times$ reduction of perception consumption, a slight reduction of the driving system, and a $3.9\%-8.5\%$ increase of driving range. 

Secondly, we report the metrics in Figure~\ref{figure: Safety Evaluation},~\ref{figure: Efficiency Evaluation},~\ref{figure: Comfort Evaluation}, and ~\ref{figure: Impact Evaluation}. None of the methods cause any collision in the test phase. For safety in Figure~\ref{figure: Safety Evaluation}, we observe that our framework \myfw\ has a TTC-R as small as the one achieved by Auto with the highest computational consumption. In addition, a smaller VEL in Figure~\ref{figure: Efficiency Evaluation} indicates that \myfw\ has a slight decrease compared to Auto in driving efficiency, and the smallest AC and DEC in Figure~\ref{figure: Comfort Evaluation} and ~\ref{figure: Impact Evaluation} indicate that \myfw\ can improve comfort and reduce disturbance to traffic flow. This is due to our robust decision module, which includes a regularization term to constrain the Q function updates in the presence of perturbed state features. It can avoid overestimating the Q values of aggressive driving actions, making the autonomous vehicle drive more smoothly.
The poor performance of Sparse, Quantize, Distill, and EcoFusion is caused by their unstable difficulty levels of the perception results and the lack of constraints on the Q function updates.
To sum up, although our framework \myfw\ can achieve a substantial decrease in computational consumption, \myfw\ can still maintain high safety as Auto and achieve the best performance in comfort and impact, with a slight decrease in traffic efficiency.

Overall, the above results indicate that our framework yields substantial energy savings across all traffic densities while preserving strong driving performance.

\subsection{Evaluation of Adaptive Perception Module}
\label{Evaluation of Adaptive Perception Module}
The adaptive perception module is used to reduce the energy consumption of perception computing. To achieve this, we first propose a classification model to distinguish perception difficulty levels of different traffic scenarios and then design a transferable tuning paradigm to explore energy-efficient knob configurations under different accuracy requirements.


\noindent \textbf{Evaluation of Scenario Classification Model.}
Firstly, our classification model is based on the image encoder Swin-T~\cite{liu2021swin}, which is a hierarchical transformer that combines local attention with shifted windows. To evidence its effectiveness and efficiency in our task, we compare it with the following image encoders: (1) ViT~\cite{yang2022maniqa}. Vision transformer that leverages global attention for superior context understanding on image patches. (2) MobileNet~\cite{sandler2018mobilenetv2}. Lightweight convolution network that uses depth-wise separable convolutions to reduce computation. (3) ResNet50~\cite{he2016deep}. Convolution network that introduces residual connections for deep networks.
Then, we design two metrics as follows: (1) Ene-Q. Average power consumption of scenario classification, which is calculated in the same way as Ene-P of the perception system. (2) Accuracy. The percentage of correctly classified instances out of the total number of instances.

We report these metrics of our method and the baselines in Table~\ref{table of accuracy of classification model}. As shown, ViT has the highest accuracy but requires a very large computational consumption. MobileNet is the most lightweight model but has a significant drop in accuracy. ResNet50 and our method with Swin-T have nearly the same computational consumption, but the accuracy of our method is much higher than the one of ResNet50. Overall, our method achieves the best balance between computational consumption and accuracy for scenario difficulty classification.

Besides a classification result, we also output an uncertainty value by adopting Monte Carlo dropout in the classification head. When the uncertainty value of a classification result is greater than 0.02, we regard this result as "uncertain" and adjust it to the highest difficulty level. 
In Table~\ref{table of uncertainty of classification model}, we present the percentage of overestimated, underestimated, and correct classification results among the uncertain results. As shown, 35.6\% of the uncertain results are overestimated, 52.1\% are underestimated, and 12.3\% are correct. Therefore, our uncertainty estimation operation is effective in finding uncertain classification results.

\begin{table}[tp]
\caption{Evaluation of Energy Consumption and Accuracy of Scenario Difficulty Classification Model}
\footnotesize
    \centering
    \begin{tabular}{|p{1.4cm}<{\centering}|p{1.8cm}<{\centering}|p{1.4cm}<{\centering}|}
         \hline
        \tabincell{c}{Methods} & \tabincell{c}{Ene-Q \\($kwh/100km$)} & \tabincell{c}{Accuracy \\ (\%)}\\
         \hline \hline
         ViT & 0.132 & \textbf{95.2}\\\hline
        MobileNet & \textbf{0.008} & 86.7\\\hline
         ResNet50 & 0.054 & 91.3\\\hline
         Swin-T & 0.058 &  94.9\\\hline
         \end{tabular}
\label{table of accuracy of classification model}
\vspace{-2mm}
\end{table}




\begin{figure}[!t]
\centering
\begin{minipage}[t]{0.4\linewidth}
\centering
\setlength{\abovecaptionskip}{0.1cm}
\includegraphics[width=\linewidth]{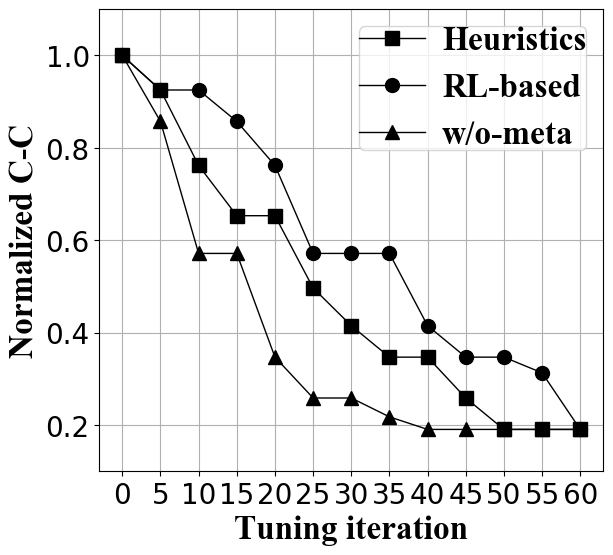}
\caption{Evaluation of Tuning Efficiency}  
\label{Tuning efficiency figure}
\end{minipage}
\hspace{2mm}
\begin{minipage}[t]{0.4\linewidth}
\centering
\setlength{\abovecaptionskip}{0.1cm}
\includegraphics[width=\linewidth]{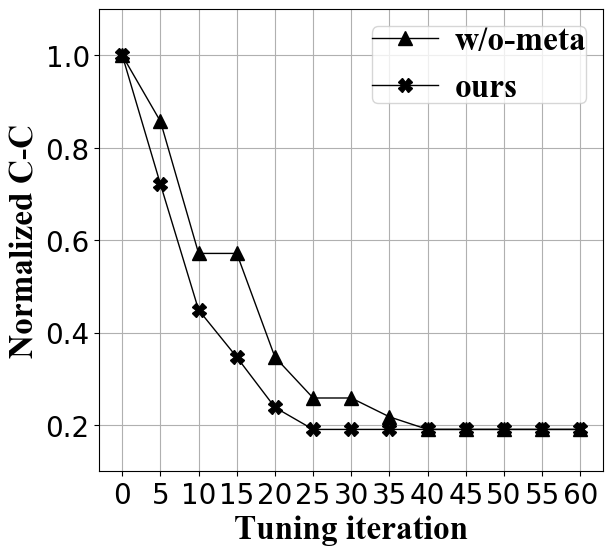}
\caption{Transferablity Analysis}
\label{Transferablity Analysis figure}
\end{minipage}
\vspace{-2mm}
\end{figure}

\noindent \textbf{Evaluation of Knob Tuning Method.}
To explore promising configurations, we use Bayesian optimization as the tuning paradigm and propose a meta-surrogate model to transfer tuning knowledge between different tuning instances. 
To evaluate the tuning performance, we compared it with two tuning methods and a variant of our method as follows: 
(1) Heuristic~\cite{zhang2017live}. A greedy local search method that searches from multiple random configurations and evaluates neighboring configurations to find a better one. 
(2) RL-based~\cite{li2019qtune}. A reinforcement learning-based tuning method that uses neural networks to approximate the configuration-accuracy relationship and select the next configuration to evaluate. 
(3) w/o-meta. A variant that removes the meta-surrogate model. Instead, it starts tuning from scratch using an initial surrogate model. To evaluate their tuning performance, we record the normalized computational consumption (named Normalized C-C) of the current optimal configurations under multiple accuracy requirements (detailed in Section~\ref{Knob tuning for Finding Promising Configurations}) and average them.

In Figure~\ref{Tuning efficiency figure}, we plot the tuning curves of Heuristic, RL-based, w/o-meta within 60 tuning iterations. All methods start tuning from scratch. As shown, w/o-meta reaches the optimal value fastest compared to Heuristic and RL-based, demonstrating the effectiveness of Bayesian optimization in our tuning task. In addition, to evidence the effectiveness of our meta-surrogate model, we compare the tuning performance of w/o-meta and our complete method in Figure~\ref{Transferablity Analysis figure}. Their tuning curves are based on the second tuning instance. In this study, there are $k$ tuning instances corresponding to $k$ difficulty levels. If using our complete method with the meta-surrogate model, the second instance can transfer tuning knowledge from the first instance. As shown, our complete method can reach the optimal value after 25 tuning iterations, fewer than the 40 iterations of w/o-meta, proving that the meta-surrogate model is effective in transferring tuning knowledge and thus speeds up the tuning process.

\begin{figure}[!t]
    \centering
    \setlength{\abovecaptionskip}{0.2cm}
    \includegraphics[width=0.5\textwidth]{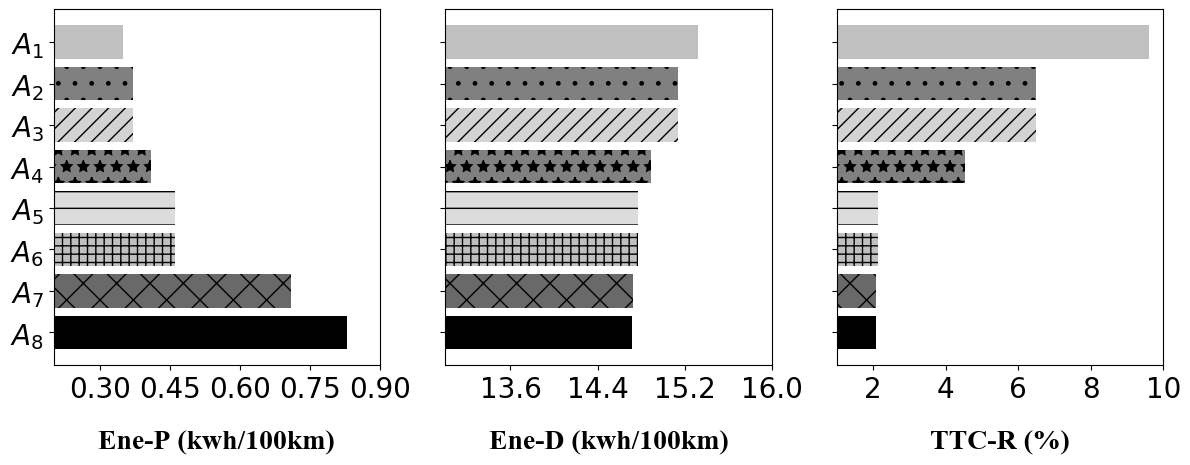}
    \caption{Effect of Accuracy Requirements}
    \label{figure: Effect of Accuracy Requirements}
\end{figure}

\noindent \textbf{Effect of Accuracy Requirements.}
Our knob tuning model can search for promising configurations under multiple accuracy requirements, i.e., $\mathcal{A}_i,\ i=1,2,3,4,5,6,7,8$, with a NDS value of 0.69, 0.70, 0.71, 0.72, 0.73, 0.74, 0.75, 0.76, respectively. After the tuning converges, we retrieve the final configuration settings in the configuration dictionary and present their computational consumption of the perception system (Ene-P) and the driving system (Ene-D), and safety metric (TTC-R) in Figure~\ref{figure: Effect of Accuracy Requirements}. As shown, a lower accuracy requirement results in lower computational consumption of the perception system, while the consumption of the driving system increases and the safety deteriorates. Specifically, the high accuracies $\mathcal{A}_7$ and $\mathcal{A}_8$ significantly increase perception computation but offer limited optimization for other metrics. The low accuracies $\mathcal{A}_1-\mathcal{A}_4$ drastically reduce the stability and safety of autonomous driving, while $\mathcal{A}_5$ and $\mathcal{A}_6$ can achieve a well-balanced performance across all metrics. 
Unless otherwise specified, the results in other experiments are based on the accuracy requirement $\mathcal{A}_6=0.74$.



\noindent \textbf{Configuration Setting at Different Difficulty Levels.}
In Table~\ref{table of used configurations}, we present the knob values of configurations in different perception difficulty levels after tuning. In this study, we set four levels (i.e., 1, 2, 3, 4), and a higher difficulty level corresponding to a more challenging traffic scenario (e.g., adverse weather conditions). 
As shown in the table, for high difficulty levels, knob values with high computation are required to achieve a desired accuracy as much as possible. Conversely, for low difficulty levels, knob values with low computation are sufficient to meet the accuracy requirement. 
In addition, we find that only scenarios with levels 1, 2, and 3 have reduced configurations, while scenarios with level 4 still maintain the highest-computation configuration. This is because even with the highest-computation configuration, these scenarios still only achieve an accuracy of 0.72, failing to meet our accuracy requirement of 0.74.
To address it, more research breakthroughs in autonomous driving perception models are needed in the future.

\noindent \textbf{Evaluation of Configuration Switching.}
In our experiments, the system switches configurations an average of 2.3 times per kilometer. However, in real-world driving scenarios, this frequency is expected to be lower, since real environments typically feature larger, more continuous scenes (e.g., long highways or urban blocks) and fewer abrupt transitions.
Since all models are resident in memory, switching between them incurs negligible latency, enabling adaptive perception with seamless model transitions.



\begin{table}[tp]
\caption{Configuration Setting at Different Difficulty Level}
\footnotesize
    \centering
    \begin{tabular}{|p{1.2cm}<{\centering}|p{1.9cm}<{\centering}|p{1.9cm}<{\centering}|p{1.9cm}<{\centering}|p{1.6cm}<{\centering}|p{1.6cm}<{\centering}|p{1cm}<{\centering}|}
         \hline
         \tabincell{c}{Difficulty \\ levels} & \tabincell{c}{$\mathit{knob}_1$} & \tabincell{c}{$\mathit{knob}_2$} & \tabincell{c}{$\mathit{knob}_3$} \\
         \hline \hline
         1 & $SparseBev$ & 5 & $prediction$\\\hline
         2 & $SparseFusion$ & 2 & $linear$ \\\hline
        3& $SparseFusion$ & 0 & \diagbox[innerwidth=1.9cm]{}{}   \\\hline
        4 & $\footnotesize BevFusion\mathord{-}e$ & 0 &  \diagbox[innerwidth=1.9cm]{}{}   \\\hline
         \end{tabular}
\label{table of used configurations}
\end{table}

\begin{table}[tp]
\caption{Evaluation of Reinforcement Leaning Models}
\footnotesize
    \centering
    \begin{tabular}{|p{1.6cm}<{\centering}|p{1cm}<{\centering}|p{1cm}<{\centering}|p{1cm}<{\centering}|p{1cm}<{\centering}|}
         \hline
         \tabincell{c}{Methods} & \tabincell{c}{TTC-R \\ ($\%$)} & \tabincell{c}{VEL \\ ($m/s$)} & \tabincell{c}{AC \\ ($m/s^2$)}
         & \tabincell{c}{DEC \\ ($m/s$)}\\
         \hline \hline
         P-DDPG & 6.42 & 18.08 & 1.45 & 0.64\\\hline
         P-DQN & 4.37 & 18.17 & 1.29 & 0.58\\\hline
         RBP-DQN & 3.45 & \textbf{18.31} & 1.14 & 0.52\\\hline
         \myfw & \textbf{2.13} & 18.38 & \textbf{0.92} & \textbf{0.44}\\\hline
         \end{tabular}
\label{Evaluation of Reinforcement Leaning Models}
\end{table} 

\subsection{Evaluation of Robust Decision Module}
In our framework, the robust decision module is used to make driving decisions based on the perturbed perception results from our adaptive perception module. It proposes a reinforcement learning model to generate driving actions with a parameterized action structure and a regularization term.

\noindent \textbf{Evaluation of Reinforcement Learning Models}
We compared our method with three reinforcement learning methods as follows:
(1) P-DDPG~\cite{hausknecht2015deep}. Deep deterministic policy gradient method that collapses the parameterized action structure into a continuous one. 
(2) P-DQN~\cite{xiong2018parametrized}. Parameterized deep Q-learning method that directly learns the parameterized action structure,
(3) RP-DQN~\cite{xia2024parameterized}. An improved P-DQN model that only restrains the update of the actor network. 
Then, we show the driving performance (i.e., safety metric TTC-R, efficiency metric VEL, comfort metric AC, impact metric DEC) of these methods under all traffic densities in Table~\ref{Evaluation of Reinforcement Leaning Models}. As shown, our method has the lowest TTC-R, AC, and DEC, demonstrating it achieves the best performance in safety, comfort, and impact compared to other methods. However, the VEL of our method has a slight decrease compared to RBP-DQN. The results indicate that our reinforcement learning can make the autonomous vehicle more cautious when facing perturbed state features, leading to larger distances from other vehicles and smoother velocity changes. The performance of RBP-DQN arises from the lack of regularization in updating Q values, leading to the misestimation of Q values of some aggressive driving actions. In addition, the poor performance of P-DDPG and P-DQN is because their model architectures struggle to optimize the action policy stably.




%% file: tex/conclusion.tex
\section{Conclusion}
In this study, we propose an energy-efficient framework \myfw\ based on modular autonomous driving, which can reduce energy consumption while maintaining good driving performance. 
In the adaptive perception module, a perception
optimization strategy is designed from the perspective of data
management and tuning. Firstly, we set the perception model, framerate, and interpolation method as knobs of the perception system. Then, we design a transferable tuning method based on Bayesian optimization to identify promising knob values that achieve low computation while maintaining desired accuracy. To adaptively switch the knob values in various traffic scenarios, a lightweight and reliable classification model is proposed to distinguish the perception difficulty in different scenarios. 
In the robust decision module, we propose a decision-making model based on reinforcement learning and design a regularization term to enhance driving stability in the face of perturbed perception results. 
Experiments on real-world and synthetic datasets evidence the superiority of our framework in both energy consumption and driving performance.
\label{Conclusion}


%% file: tex/ack.tex
\section{Acknowledgment}
This work is partially supported by NSFC (No. 62472068, 62272086), and Municipal Government of Quzhou under Grant (No. 2024D036, 2024D037).